\documentclass[11pt,a4paper]{article}
\usepackage[hyperref]{acl2020}
\usepackage{times}
\usepackage{latexsym}

\usepackage[utf8]{inputenc}
\usepackage[english]{babel}
\usepackage{verbatim}
\usepackage{pgfplots}
\usepackage{graphicx}
\pgfplotsset{width=7cm,compat=1.8}
\usepackage{ragged2e}

\usepackage{microtype}

\aclfinalcopy 
\title{LCP-RIT at SemEval-2021 Task 1: Exploring Linguistic Features for Lexical Complexity Prediction}

\author{Abhinandan Desai, Kai North, Marcos Zampieri, Christopher M. Homan \\
  Rochester Institute of Technology \\
  Rochester, NY, USA \\
  \texttt{ad2724@rit.edu, kn1473@rit.edu
  mazgla@rit.edu, cmh@cs.rit.edu} \\
}

\date{}

\begin{document}

\maketitle

\begin{abstract}
This paper describes team LCP-RIT's submission to the SemEval-2021 Task 1: Lexical Complexity Prediction (LCP). The task organizers provided participants with an augmented version of CompLex \cite{shardlow-etal-2020-complex}, an English multi-domain dataset in which words in context were annotated with respect to their complexity using a five point Likert scale. Our system uses logistic regression and a wide range of linguistic features (e.g. psycholinguistic features, $n$-grams, word frequency, POS tags) to predict the complexity of single words in this dataset. We analyze the impact of different linguistic features in the classification performance and we evaluate the results in terms of mean absolute error, mean squared error, Pearson correlation, and Spearman correlation.
\end{abstract}

\section{Introduction}\label{intro}

Lexical complexity prediction (LCP) is the task of predicting the complexity value of a target word within a given text \cite{shardlow-etal-2020-complex}. Complexity within LCP is used as a “synonym for difficulty" \cite{malmasi-zampieri:2016:SemEval}\footnote{The term “complex" within LCP is not necessarily related to the terms \emph{simplex} and \emph{complex} used in morphology.}. A complex word is therefore a word that a target population may find difficult to understand. Various LCP systems have been designed to identify words that may be found to be complex for children \cite{kajiwara2013selecting}, language learners \cite{malmasi-dras-zampieri:2016:SemEval}, or people suffering from a reading disability, such as dyslexia \cite{Rello2013}. These systems have been utilized within assistive language technologies, lexical simplification systems, and in a variety of other applications.

LCP is related to complex word identification (CWI) \cite{paetzold-specia:2016:SemEval1}. CWI is modeled as a binary classification task by assigning each target word with a complex or non-complex label. The shortcomings of modeling lexical complexity using binary labels have been discussed in previous work \cite{zampieri-EtAl:2017:NLPTEA,maddela2018word}, motivating the organization of SemEval-2021 Task 1: Lexical Complexity Prediction.\footnote{\url{https://sites.google.com/view/lcpsharedtask2021/home}} LCP models complexity in a continuum and the goal is to predict a target word's degree of complexity by assigning it a value between 0 and 1. This value may then correspond to one of the following labels: very easy (0), easy (0-0.25), neutral (0.25-0.5), difficult (0.5-0.75), or very difficult (0.75-1) \cite{shardlow-etal-2020-complex}.

In this paper, we describe (in detail in Section \ref{system}) the LCP-RIT entry to SemEval-2021 Task 1. We approached LCP from a feature engineering perspective with a particular focus on the adoption of psycholinguistic features, such as average age-of-acquisition (AoA), familiarity, prevalence, concreteness, and arousal, alongside the use of prior complexity labels. Our submitted system utilized a combination of these linguistic features, which we compared to a baseline model that only used statistical features: word length, word frequency and syllable count \cite{quijada-medero:2016:SemEval, mukherjee-EtAl:2016:SemEval}. On our training dataset, our submitted system achieved a mean absolute error (MAE) of 0.067, mean squared error (MSE) of 0.007, Person Correlation (R) score of 0.779, and a Spearman Correlation ($\rho$) score of 0.724. This surpassed our baseline model's performance by a MAE of 0.008, MSE of 0.003, as well as R and $\rho$ scores of 0.075 and 0.062 respectively.

\section{Related Work}\label{background}

Before SemEval-2021 Task 1: LCP, two CWI shared tasks were organized at one SemEval-2016 and the other at BEA-2018 \cite{paetzold-specia:2016:SemEval1, stajner-EtAl:2018:BEA}. While the first CWI provided participants with an English dataset, the second provided a multilingual dataset. The systems submitted to the English track of the second shared task \cite{stajner-EtAl:2018:BEA} performed better overall than the previous task \cite{paetzold-specia:2016:SemEval1}, probably due to the properties of the two datasets \cite{zampieri-EtAl:2017:NLPTEA}. State-of-the-art neural net models and word embedding models performed worse than conventional models such as decision trees (DTs) and random forests (RFs) \cite{stajner-EtAl:2018:BEA}. Among the conventional models, the use of statistical, character $n$-gram, and psycholinguistic features was found to be highly effective in improving CWI performance \cite{malmasi-dras-zampieri:2016:SemEval,zampieri-tan-vangenabith:2016:SemEval,paetzold-specia:2016:SemEval1,stajner-EtAl:2018:BEA}.

Among the best performing systems in CWI 2018, \citet{syspaper7} used an ensemble of classifiers. They found that during their system's development, the boosting classifier AdaBoost, a random forest classifier, or a combination of both classifiers achieved the highest performance. These systems used multiple features such as the word's grammatical category, Google character n-gram frequency as well as a range of psycholinguistic features \cite{syspaper7}. 

Of the remaining systems, \citet{syspaper11} and \citet{syspaper4} utilized statistical features, such as word length and number of syllables, psycholinguistic features such as familiarity, age of acquisition, concreteness, and imagery scores, and word $n$-grams. \citet{syspaper4} compared the performance of tree ensembles to a convolutional neural network (CNN). They found that their tree ensembles performed better than their CNN, especially when the target expression contained more than three words \cite{syspaper11}. 

\section{Task and Dataset}\label{task}

The LCP shared task organizers provided participants with the CompLex corpus, an English multi-domain dataset with sentences from the Bible, the European Parliament proceedings, and a collection of biomedical texts. A pool of annotators, using a five point Likert scale, labeled the complexity of single words and multi-word expressions in CompLex \cite{shardlow-etal-2020-complex}.

Taking advantage of the annotation of single words and multi-word expressions, the LCP shared task was divided into two sub-tasks as follows:

\begin{itemize}
    \item {\bf Sub-task 1:} predicting the complexity score for single words; 
    \item {\bf Sub-task 2:} predicting the complexity score for multi-word expressions.
\end{itemize}

\noindent We chose to participate in sub-task 1. Sub-Task 1's training dataset contained 7,662 instances with its test dataset having 917 instances. 20\% of the training dataset was used to test our system's performance during development. Sub-Task 1 received 54 system submissions. 

\section{System Overview}\label{system}

\subsection{Model}\label{model}

Taking inspiration from the CWI systems discussed in Section \ref{background}, we adopted a random forest regressor (RFR) to predict the complexity values of each word within the test dataset. To achieve this, we tested the impact a variety of linguistic features have on LCP performance during our system's development. The RFR was taken from scikit-learn's ensemble module \cite{scikit-learn}. The RFR used a maximum of 120 trees and 750 features.

\subsection{Features}\label{features}

We constructed a baseline RFR using the following statistical features and character trigrams. We then used psycholinguistic and additional features to see whether its baseline performance could be improved.

\textbf{Statistical Features} include word length, word frequency and syllable count. Zipf's Law implies that words that appear less frequently within a text are likely to be longer and therefore may be considered more complex than words that are more frequent and shorter \cite{quijada-medero:2016:SemEval}. In addition, words with a high number of syllables are difficult to pronounce and are subsequently hard to read \cite{mukherjee-EtAl:2016:SemEval}. As such, word length, word frequency and syllable count were considered to be good baseline statistical indicators of a word's complexity value.
  
\textbf{Character N-grams} include the use of character bigram and trigram frequencies. These frequencies were calculated by counting each bigram's and trigram's presence in the target words provided in Sub-Task 1's training dataset. Experimentation with bigrams and trigrams, along with a combination of both, found that the use of trigrams on their own was superior. This together with their use within prior CWI systems justified their inclusion within our baseline model \cite{stajner-EtAl:2018:BEA}. 
  
\textbf{Psycholinguistic Features} include average age of acquisition (AoA), concreteness, familiarity, prevalence and arousal. AoA is the age at which a word's meaning is first learned. Concreteness refers to “the degree to which the concept denoted by a word refers to a perceptible entity” \cite{Brysbaert2013}. Familiarity and prevalence are somewhat similar. Familiarity is how well known the word is to an individual and was obtained through self-report \cite{gilhooly1980}. Prevalence was calculated in accordance to the percentage of people who knew the word \cite{Brysbaertetal2019}. Lastly, arousal is a measure of how active or passive a word's meaning is interpreted as being\footnote{The terms “active" and “passive" do not refer to the use of \emph{active} or \emph{passive voice} but rather the emotional or physical intensity associated with a word's meaning \cite{Mohammad2018}.}. For instance, the word “\emph{nervous}” indicates more arousal than “\emph{lazy}” \cite{Mohammad2018}. As such, grammatical categories such as adjectives, verbs, and adverbs may incite higher levels of arousal than nouns. 
  
Average AoA was calculated by averaging the AoAs provided in the Living Word Vocabulary Dataset \cite{DaleORourke1981} with an updated version of this dataset \cite{BrysbaertBiemiller2017}. Both datasets consisted of AoA values for 44,000 English word meanings. Concreteness, familiarity and arousal values were taken from the MRC Psycholinguistic Database \cite{wilson1988mrc} as well as three newer datasets each containing 37,058, 61,858 and 20,000 English words \cite{Brysbaert2013, Brysbaertetal2019, Mohammad2018}.
  
\textbf{Additional Features} include part-of-speech (POS) tags as well as prior complexity labels. POS tags were generated by using the Python Natural Language Toolkit \cite{bird2009natural}. Prior complexity labels were taken from the previous CWI shared tasks \cite{paetzold-specia:2016:SemEval1, stajner-EtAl:2018:BEA} and the Word Complexity Lexicon \cite{maddela2018word}. A combined dataset was then created that contained a total of 26,088 English words each with a binary complexity value.
  
\section{Evaluation}\label{evaluation}

\subsection{Features}

To determine the effect each feature had on our baseline model's performance, we used the following scores: mean absolute error (MAE), mean squared error (MSE), Pearson Correlation (R) and Spearman Correlation ($\rho$). Table 1 depicts each feature's performance on the training dataset. These criteria have also been used in the SemEval LCP test set evaluation. 

Average AoA decreased the baseline model's MAE and MSE by 0.004 and 0.001 respectively. It likewise increased its R and $\rho$ scores by 0.039. This generated the second highest R and $\rho$ scores of 0.743 and 0.701 respectively. Average AoA is therefore a useful feature for LCP. 

Brysbaert et al.'s prevalence and concreteness \cite{Brysbaert2013, Brysbaertetal2019} were also seen to improve the baseline model's performance with prevalence being the most notable. Prevalence decreased baseline MAE and MSE scores by 0.005 and 0.002 respectively. It also surpassed baseline R by 0.054 and $\rho$ by 0.047, yielding the highest increases among all features. Concreteness \cite{Brysbaert2013} also caused a slight increase in the baseline model's scores, being greater than that caused by MRC concreteness. Concreteness values \cite{Brysbaert2013} increased the baseline model's R and $\rho$ scores by 0.032 and 0.024 respectively, whereas the MRC concreteness values resulted in a slightly less impressive increase of 0.019 in both its R and $\rho$ scores. However, there was little-to-no difference in MAE and MSE produced by either set of concreteness values.

\begin{table}[!ht]
\scalebox{0.90}{
\begin{tabular}{lcccc}
\hline
\multicolumn{1}{c}{} & \multicolumn{4}{c}{\textbf{Performance}}\\
    \hline
     \textbf{Features} & R & $\rho$ & MAE & MSE\\
     \hline
     Baseline Features & 0.704 & 0.662 & 0.075 & 0.010\\
     Average AoAs & 0.743 & 0.701 & 0.071 & 0.009\\
     \textbf{Prevalence} & \textbf{0.758} & \textbf{0.709} & \textbf{0.070} & \textbf{0.008}\\
     MRC Familiarity & 0.727 & 0.687 & 0.073 & 0.009\\
     Concreteness & 0.736 & 0.686 & 0.072 & 0.009\\
     MRC Concreteness & 0.723 & 0.681 & 0.073 & 0.009\\
     Arousal & 0.722 & 0.676 & 0.074 & 0.009\\
     POS Tags & 0.701 & 0.663 & 0.075 & 0.010\\
     Complexity Labels & 0.727 & 0.686 & 0.072 & 0.009\\
     \hline     
\end{tabular}
}
 \caption{\label{tab:table-name}Feature performance on training dataset. The baseline model uses the statistical features and character trigrams. Best results in bold.}
\end{table}

\noindent Two possible conclusions can be drawn: 1). The difference in the calculation of prevalence versus that of familiarity likely causes prevalence to be a greater indicator of word complexity\footnote{Prevalence being the percentage of people who. know the word \cite{Brysbaertetal2019}. Familiarity being a self-reported measure of an individual's awareness of the word \cite{gilhooly1980}.}, and 2). The superior coverage of Brysbeart et al.'s prevalence (2018) and concreteness (2013) datasets \cite{Brysbaertetal2019, Brysbaert2013}: being 52.62\% and 57.51\% respectively, compared to that of the MRC Psycholinguistic Database \cite{wilson1988mrc}: being 23.44\%, suggests that there now exists larger and more up-to-date psycholinguistic datasets that are more useful for LCP feature engineering. 

Arousal has never before been used for LCP or CWI. Due to its ability to differentiate grammatical categories, such as nouns and verbs, along with its ability to signify a word's intensity, we had speculated that arousal would be able to help predict a word's complexity. Arousal was found to have no significant effect on the baseline model's performance. Nevertheless, once added to our submitted system, it slightly decreased its MSE by 0.001 and increased its R and $\rho$ scores by 0.002 and 0.001 respectively. 

POS tags was the worst performing feature as POS tags had little affect on improving our model's performance. It achieved the same MAE and MSE values as our baseline model: 0.075 and 0.01 respectively. Regarding R score, only a slight increase of 0.001 was observed. POS tags was the only feature that saw a decrease in our model's $\rho$ score, worsening its performance by 0.003. This suggests that a word's grammatical category may not impact its degree of complexity. This is also supported by Arousal's lack of improved performance.    

Given that prior complexity labels are directly related to complexity prediction, it was believed that they would be the most influential in improving overall performance. Instead, AoA, prevalence and concreteness were all found to be more beneficial with higher or identical MAE, MSE, $\rho$, and R scores. Complexity labels only saw a slight decrease in MAE and MSE by 0.003 and 0.001 respectively and a slight increase in $\rho$ and R scores by 0.023 and 0.024 respectively. The binary nature of prior CWI datasets is likely responsible for this phenomenon, as binary 0 or 1 complexity values are not well suited for a regression-based task, such as LCP. This would have resulted in the same problem faced by previous CWI systems: the misclassification of words on the decision boundary.

\subsection{Models}

The results for the three models on the training dataset are presented in Table \ref{tab:resultsmodels}. This is then followed by a short description of each model as well as our performance on the test dataset.

\begin{table}[!ht]
\centering
\scalebox{0.95}{
\begin{tabular}{lccccc}
\hline
\multicolumn{1}{c}{} & \multicolumn{4}{c}{\textbf{Performance}}{}\\
    \hline
     \textbf{Model} & R & $\rho$ & MAE & MSE\\
     \hline
     Model 1 & 0.772 & 0.717 & 0.068 & 0.008\\
     Model 2 & 0.777 & 0.724 & 0.067 & 0.008\\
     LCP-RIT & \textbf{0.779} & \textbf{0.724} & \textbf{0.067} & \textbf{0.007}\\
     \hline     
\end{tabular}
}
 \caption{\label{tab:resultsmodels}Model performance on training dataset.}
\end{table}

\paragraph{Model 1 - Top 3 Features:} Adding the top 3 features of average AoA, prevalence and concreteness to our baseline model reduced its MAE and MSE by 0.007 and 0.002 respectively and increased its R score by 0.068 and its $\rho$ score by 0.055. It attained a new MAE of 0.068 which was noticeably better than our baseline model's previous MAE of 0.075. This goes to the show that inclusion of psycholinguistic features has a positive impact on the performance of an LCP system. 

\paragraph{Model 2 - Top 5 Features:}\label{results3}

A small improvement was seen after adding the fourth and fifth best performing features to Model 1, namely, MRC familiarity and prior complexity labels. Model 2 increased Model 1's R and $\rho$ scores by 0.005 and 0.007. However, it failed to improve Model 1's MAE or MSE. This small increase in performance was due to the prior top 3 features of average AoA, prevalence and concreteness already having captured those instances caught by MRC familiarity and prior complexity labels. This further proves the redundancy of the MRC Psycholinguistic Database \cite{wilson1988mrc} as well as binary complexity labels for LCP feature engineering.

\paragraph{LCP-RIT:}\label{results4}

Our final model submitted to the official evaluation used the psycholinguistic features of average AoA, prevalence, concreteness and arousal together with our baseline model's features of word length, syllable count, word frequency and character trigrams to predict the lexical complexity of single words. On the training dataset of SemEval-2021 Task 1: LCP, we achieved a MAE of 0.067, MSE of 0.007, R score of 0.779, and $\rho$ score 0.724. We performed less well on the single word test dataset with an MAE and MSE of 0.072 and 0.009 respectively and $\rho$ and R scores of 0.709 and 0.653 respectively. This reduced performance may be indicative of our submitted system being overfit on our training dataset. 

\section{Conclusion}\label{results}

We carried out multiple experiments evaluating the impact of linguistics features in LCP using the CompLex dataset for English. We have shown that several psycholinguistic features help with LCP. Average AoA, prevalence and concreteness were all found to be beneficial, whereas MRC familiarly, MRC concreteness and prior complexity labels were proven to be redundant. We would like to explore other features described in \newcite{shardlow2021predicting}. In terms of performance, we believe that the multiple features we tested allowed us to get close to the maximum performance for this dataset using regression. A possible alternative for better performance is to test state-of-the-art transformer models. Furthermore, we are interested in looking at the performance of these features for LCP in languages other than English and for multilingual datasets. 

\section*{Acknowledgments}

We would like to thank the LCP shared task organizers for proposing this interesting shared task and for making the data available. 

\bibliography{CWI}
\bibliographystyle{acl_natbib}

\end{document}